\documentclass{article}

\usepackage[preprint]{neurips_2025}

\usepackage[utf8]{inputenc}
\usepackage[T1]{fontenc}
\usepackage{hyperref}
\usepackage{url}
\usepackage{booktabs}
\usepackage{amsfonts}
\usepackage{nicefrac}
\usepackage{microtype}
\usepackage{amsmath}
\usepackage{amssymb}
\usepackage{bbm}
\usepackage[dvipsnames]{xcolor}
\usepackage{graphicx} 
\usepackage{subcaption}
\usepackage{adjustbox}  
\usepackage{float}
\usepackage{booktabs}
\usepackage{tabularx}

\newcolumntype{C}{>{\centering\arraybackslash}X}
\newcolumntype{R}{>{\raggedleft\arraybackslash}X}

\newcommand{\pms}{\textnormal{{B-SiLU}}}
\newcommand{\nelu}{\textnormal{{NeLU}}}

\newcommand{\ra}[1]{\renewcommand{\arraystretch}{#1}}

\newcommand{\best}[1]{%
\textbf{\textcolor{magenta}{\boldmath #1}}
}
\newcommand{\secondbest}[1]{%
\textbf{\textcolor{PineGreen}{\boldmath #1}}
}

\title{The Resurrection of the ReLU}

\author{%
  Coşku Can Horuz$^{1}$\thanks{Equal contribution, corresponding authors.}
  \quad
  Geoffrey Kasenbacher$^{1,2}$\footnotemark[1]
  \quad
  Saya Higuchi$^{1}$ 
  \quad
  Sebastian Kairat$^{1}$\\
  \textbf{Jendrik Stoltz}$^{1}$ 
  \quad
  \textbf{Moritz Pesl}$^{1}$ 
  \quad
  \textbf{Bernhard A. Moser}$^{3,4}$ 
  \quad
  \textbf{Christoph Linse}$^{5}$\\
  \textbf{Thomas Martinetz}$^{5}$
  \quad
  \textbf{Sebastian Otte}$^{1}$
  \vspace{0.2cm}\\
  $^{1}$Institute of Robotics and Cognitive Systems, University of Lübeck \\
  $^{2}$Mercedes-Benz AG \\
  $^{3}$Software Competence Center Hagenberg (SCCH) \\
  $^{4}$Institute of Signal Processing, Johannes Kepler University of Linz (JKU) \\
  $^{5}$Institute of Neuro- and Bioinformatics, University of Lübeck\vspace{0.2cm}\\
  \small
  \texttt{\{cosku.horuz, sa.higuchi, sebastian.kairat,}\\
  \small
  \texttt{c.linse, thomas.martinetz, sebastian.otte\}@uni-luebeck.de}\\
  \small
  \texttt{\{jendrik.stoltz, moritz.pesl\}@student.uni-luebeck.de}\\
  \small
  \texttt{geoffrey.kasenbacher@mercedes-benz.com}\\
  \small
  \texttt{bernhard.moser@scch.at}
}

\begin{document}

\maketitle

\begin{abstract}
Modeling sophisticated activation functions within deep learning architectures has evolved into a distinct research direction. Functions such as GELU, SELU, and SiLU offer smooth gradients and improved convergence properties, making them popular choices in state-of-the-art models. Despite this trend, the classical ReLU remains appealing due to its simplicity, inherent sparsity, and other advantageous topological characteristics. However, ReLU units are prone to becoming irreversibly inactive---a phenomenon known as the dying ReLU problem---which limits their overall effectiveness. In this work, we introduce surrogate gradient learning  for ReLU (SUGAR) as a novel, plug-and-play regularizer for deep architectures. SUGAR preserves the standard ReLU function during the forward pass but replaces its derivative in the backward pass with a smooth surrogate that avoids zeroing out gradients.
We demonstrate that SUGAR, when paired with a well-chosen surrogate function, substantially enhances generalization performance over convolutional network architectures such as VGG-16 and ResNet-18, providing sparser activations while effectively resurrecting dead ReLUs.
Moreover, we show that even in modern architectures like Conv2NeXt and Swin Transformer---which typically employ GELU---substituting these with SUGAR yields competitive and even slightly superior performance.
These findings challenge the prevailing notion that advanced activation functions are necessary for optimal performance. Instead, they suggest that the conventional ReLU, particularly with appropriate gradient handling, can serve as a strong, versatile revived classic across a broad range of deep learning vision models.
\end{abstract}

\section{Introduction}
\label{sec:introduction}

The choice of activation functions in deep neural networks has a substantial effect on the convergence and performance of a model. Prior to the suggestion to apply Rectified Linear Unit (ReLU) as activation function in \cite{glorot2011relu}, concerted effort was made to tackle the saturation and vanishing gradient problem of sigmoidal activations. ReLU has been shown to accelerate convergence and often also to enhance generalization. As a landmark moment in deep learning history, ReLU has been used in AlexNet \cite{krizhevsky2012alexnet}, which significantly outperformed its competitors on the ImageNet \cite{deng2009imagenet} benchmark. This work explicitly studied the superiority of ReLU over hyperbolic tangent (tanh) in a deep convolution network. Since then, ReLU has been used in numerous tasks including image classification and segmentation, reinforcement learning, natural language processing, and speech recognition \cite{he2016deep, ronneberger2015unet, andrew2024relu, vaswani2017attention, jha2024relusrevival, zeiler2013rectified}. 

Mathematically, recent research has established a deep connection between tropical geometry and feedforward neural networks with ReLU activations~\cite{ZhangNL18}, unraveling an underlying algebraic structure of ReLU, namely, tropical semiring that replaces addition with it's maximum and multiplication with addition. As a result, the decision boundaries formed by ReLU networks correspond to tropical hypersurfaces, which are the fundamental objects in tropical geometry.  Such neural networks are equivalent to tropical rational functions \cite{ZhangNL18}, which is defined as a ratio of two tropical polynomials, themselves formed by maxima of affine functions. Closely related to this is the intuitive geometric understanding of consecutive layers of ReLU networks as space folding transformations that allow a compact representation of similarities in the input space~\cite{lewandowski2025on, Montufar2014, naitzat2020topology}. 

In ReLU networks, the negative pre-activations are silenced with true zeros, which allow sparse representations. This might lead to better generalization on unseen data, but forcing too much sparsity may damage the prediction as it effectively reduces the model capacity \cite{glorot2011relu}. This phenomenon, known as ``the dying ReLU problem'', has been a big caveat of ReLU networks. The dying ReLU problem led to a plethora of linear unit functions including, but not limited to LeakyReLU \cite{maas2013leakyrelu}, PReLU \cite{he2015prelu}, GELU \cite{hendrycks2023gelu}, SELU \cite{klambauer2017selu}, SiLU/Swish \cite{elfwing2018silu, ramachandran2018swish}, ELU \cite{clevert2016elu}. All these functions introduce non-zero activations for negative pre-activation values, offering different trade-offs between ReLU’s modeling benefits and the advantages of smooth, continuous gradients.

In this paper, we address the limitations of ReLU without sacrificing its advantages by introducing a novel approach: the application of a Surrogate Gradient for ReLU (SUGAR). SUGAR allows models to retain standard ReLU activations, while ensuring stable gradient flow even for negative pre-activations. Further contributions of this work are as follows:
\begin{itemize}
    \item We propose two new surrogate gradient functions, {B-SiLU} and {NeLU}, which integrate seamlessly into a variety of models. These functions consistently improve generalization performance.
    \item We conduct comprehensive experiments with VGG-16~\cite{simonyan2015very} and ResNet-18~\cite{he2015deep}, demonstrating that SUGAR significantly enhances generalization in both architectures.
    \item We evaluate SUGAR on modern architectures such as Swin Transformer~\cite{liu2021Swin} and Conv2NeXt~\cite{feng2022conv2next}, showcasing its adaptability and effectiveness.
    \item An in-depth analysis of VGG-16 layer activations reveals a distinct shift in activation distributions when SUGAR is applied, providing visual evidence for its role in mitigating the dying ReLU problem, while at the same time more sparse representations are fostered.
    \item We further explore the loss landscape with and without SUGAR, offering deeper insight into its optimization benefits.
\end{itemize}

The proposed SUGAR method offers several desirable properties. It is simple to implement and consistently utilizes ReLU in the forward pass. When combined with the proposed \pms{} surrogate function, VGG-16 achieves improvements of 10 and 16 percentage points in test accuracy on CIFAR-10 and CIFAR-100 \cite{cifarkrizhevsky2009learning}, respectively, while ResNet-18 shows corresponding gains of 9 and 7 percentage points compared to the best-performing models without SUGAR.
\section{Background}\label{sec:background}
\subsection{Surrogate gradient learning}
In conventional artificial neural networks, learning relies on gradient-based optimization methods such as backpropagation, which require continuous, differentiable activation functions. However, spiking neural networks (SNNs) are discrete and non-differentiable, making direct application of backpropagation infeasible. 

{Surrogate gradient learning} emerged as a solution to train SNNs by replacing the gradient of the non-differentiable spiking function with a smooth and differentiable approximation \cite{neftci2019surrogate}. These surrogate functions allow gradients to flow through the network during training, enabling the use of powerful optimization techniques in neuromorphic and event-based computing settings. This idea gained traction in the late-2010s and has since become a cornerstone for training biologically inspired spiking models in a computationally efficient manner \cite{bellec2018long,Higuchi2024Balanced,yin2021accurate,zenke2021remarkable}.

Recently, a related approach called ProxyGrad \cite{linse2024proxygrad} improved activation maximization (AM) in convolutional networks by manipulating gradients during optimization. The study showed that using LeakyReLU in the backward pass, while keeping ReLU in the forward pass, enabled AM to escape poor local optima and reach higher activation values. As a result, the method produced more informative and interpretable feature visualizations.

\subsection{Forward gradient injection (FGI)}
Forward Gradient Injection (FGI) is the backbone algorithm in SUGAR. It was introduced in \cite{otte2024flexible} as a surrogate gradient strategy in SNNs. It exploits the {stop gradient operator} (i.e. \texttt{.detach()}) to manipulate the gradients such that a model with non-differentiable spikes becomes trainable via gradient signals. FGI enables a gradient injection during forward pass with the following equation (indirect surrogate gradient function):
\begin{equation}\label{eq:fgi}
    y = g(x) - \operatorname{sg}(g(x)) + \operatorname{sg}(f(x))
\end{equation}
where $\operatorname{sg}(\cdot)$ is the stop gradient operator, $f(\cdot)$ is the forward computation with its gradient bypassed and $g(\cdot)$ is another function where the gradient is instead injected over the variable of interest $x$ while not contributing to the forward result due subtraction with itself. Choosing $g(\cdot)$ as an activation function with non-zero gradients for negative inputs, ReLU networks can be trained without suffering from the dying ReLU problem. However, \autoref{eq:fgi} requires the gradient computation of $g(\cdot)$ in the backward pass. Following \textit{multiplication trick} allows $\tilde{g}(\cdot)$ to be exactly the derivative of ${f}(\cdot)$ in the backward pass (direct surrogate gradient function):
\begin{align}\label{eq:mul_fgi}
    m &= x \cdot \operatorname{sg}(\tilde{g}(x))\\
    y &= m - \operatorname{sg}(m) + \operatorname{sg}(f(x))
\end{align}

FGI enables the injection of surrogate gradient functions directly in the forward pass, independent of the original function $f(\cdot)$. The results in \cite{otte2024flexible} suggest that FGI can improve both model optimizability and exportability over classical deployments of surrogate gradients (i.e. as overriding the backward function).
\section{Surrogate gradient for ReLU (SUGAR)}
\label{sec:method}
Our proposed method applies FGI in ReLU networks with smooth surrogate functions. We extend the potential applications of surrogate gradients beyond SNNs, and aim to introduce this framework as a technique to overcome disadvantages of ordinary ReLUs. 

Indirect FGI within the context of SUGAR can be expressed as:
\begin{equation}
y = f(x) - \operatorname{sg}(f(x)) + \operatorname{sg}(\operatorname{ReLU}(x))
\end{equation}
This formulation enables gradient injection and ensures gradient propagation even for negative activations. Specifically, using the multiplication trick from \cite{otte2024flexible}, the direct injection of the surrogate gradient function is accomplished via:
\begin{align}
m &= x \cdot \operatorname{sg}(\tilde f(x)) \label{eq:sugar_mul_trick1}\\
y &= m - \operatorname{sg}(m) + \operatorname{sg}(\operatorname{ReLU}(x)) \label{eq:sugar_mul_trick2}
\end{align}
Here, $\tilde f(x)$ explicitly defines the surrogate gradient behavior for ReLU.

The choice of the surrogate function is flexible and can include activation functions commonly employed in state-of-the-art applications, such as ELU, GELU, SiLU, SELU, and Leaky ReLU (see \autoref{fig:act_functions}). These functions typically possess desirable properties motivated by mechanisms like self-gating or self-normalization. Importantly, these surrogate candidates share the characteristic of having non-zero gradients for negative inputs ($x<0$), unlike ReLU. Although surrogate functions enable gradient flow for negative activations, the forward pass and subsequent loss computation strictly depend on activations for $x>0$. Consequently, the effect of surrogate gradient learning in this setting can be interpreted as filtering out pre-activations below the cut-off threshold, reducing the network's propensity to overfit due to topological simplification and sparsity \cite{mehta2019implicit, naitzat2020topology}, but without harming the gradient flow.

In preliminary studies, we realized the need for adapting the current activation functions to utilize the specific purpose of SUGAR. As it will be discussed in depth later, SUGAR's effect varies in different regularization settings. Therefore, in the following, we propose two new surrogate functions that align well with these settings.

\subsection{B-SiLU}

We introduce a novel activation function named Bounded Sigmoid Linear Unit (\pms{}), which combines self-gating characteristics with a tunable lower bound parameter. Mathematically, this function can be expressed as:
\begin{equation}
\pms{}(x) = (x + \alpha) \cdot \sigma(x) - \frac{\alpha}{2}, \quad \text{with } \alpha = 1.67
\end{equation}
where $\sigma(x)$ denotes the sigmoid activation. The derivative of the \pms{} activation function is given by:
\begin{equation}
\frac{d}{dx}\pms{}(x) = \sigma(x) + (x + \alpha) \sigma(x) (1 - \sigma(x))
\end{equation}
Both \pms{} and its derivative are visualized in \autoref{fig:act_functions}.
The \pms{} activation function emerged from exploratory experiments with SUGAR, drawing particular inspiration from the swish-like sigmoidal function introduced in \cite{rozell2008sparse} and the related thresholding operator examined in \cite{kasenbacher2025warp,zeng2014l_}.
It is motivated by the goal of combining advantageous properties from the self-gating behavior of SiLU and the smoothness of GELU.

\subsection{NeLU}
We further introduce Negative slope Linear Unit (\nelu), as a smooth derivative substitute for ReLU. It is inspired by the constant derivative of ReLU for $x > 0$ and a smooth negative slope from GELU for $x < 0$. For large negative inputs, the resulting gradient converges back to zero.

\begin{equation}\label{eq:nelu}
\frac{d}{dx}\nelu(x) = 
\begin{cases}
1, & \text{if } x > 0 \\
\displaystyle \alpha \frac{2x}{\left(1 + x^2\right)^2}, & \text{else}
\end{cases}
\end{equation}

\begin{figure}[h!]
    \centering
    \includegraphics[width=1.0\linewidth]{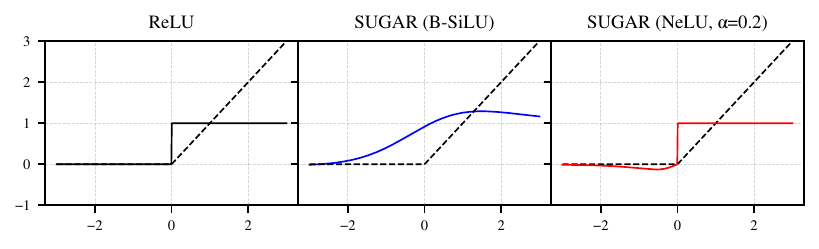}
    \caption{Comparison of activation functions and their derivatives. From left to right: ReLU (dashed) and its derivative (black), ReLU activation function with B-SiLU derivative (blue) and ReLU activation function with NeLU derivative (red).
    }
    \label{fig:activation_comparison}
\end{figure}

$\alpha$ controls the magnitude of the negative gradient for small values of $x < 0$, which ensures stability. The resulting gradient is shown in \autoref{fig:activation_comparison}. By applying the multiplication trick in \autoref{eq:sugar_mul_trick1} and \autoref{eq:sugar_mul_trick2}, it is possible to directly set the gradient of the activation as in \autoref{eq:nelu}.

\section{Experiments}\label{experiments}
\subsection{Surrogate function comparison on CIFAR-10/100}\label{subsec:surr}
We conducted extensive experiments to evaluate and compare various activation functions, both with and without SUGAR, on the CIFAR-10 and CIFAR-100 datasets using ResNet-18 and VGG-16 architectures. In every run where SUGAR is applied, the forward function is always the standard ReLU; for a given surrogate activation \(f\) (e.g.\ ELU), we compare the performance of the network using \(f\) in the forward pass and it's true gradient in the backward pass (non-SUGAR scenario) against the identical architecture using ReLU for the forward pass but \(f\)’s gradient for the backward pass (e.g. SUGAR with ELU). Each configuration was trained five times with different random seeds to ensure robustness of the results. To isolate the generalization effects of the activation functions and the application of SUGAR we did not apply any data augmentation (see \autoref{appendix:experiment_vgg_resnet} for the full experimental setup). The set of surrogate functions evaluated included LeakyReLU, SELU, ELU, GELU, SiLU (Swish), Mish, and our proposed \pms{} and \nelu. See \autoref{appdx:vgg_resnet} for the complete experimental results.

\begin{figure}[h!]
  \centering
  \includegraphics[width=5.5in]{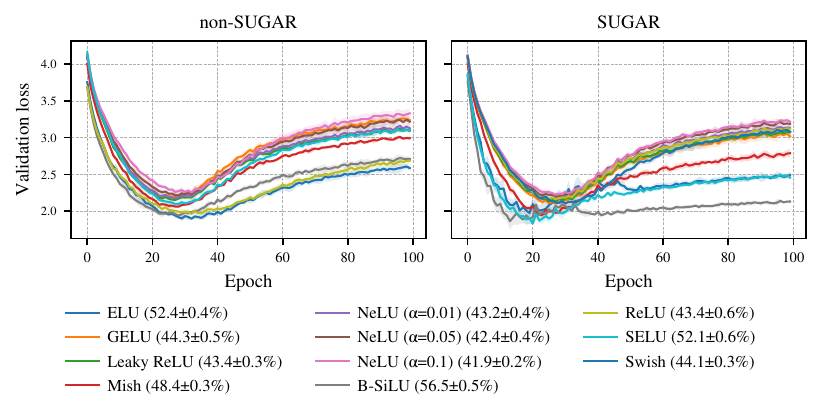}
  \caption{The plots show the validation loss of ResNet-18 on CIFAR-100 with and without SUGAR. In the legend, the corresponding test accuracies (of the respective functions as surrogates) are given for completeness. See \autoref{app:training_curves} for all the convergence plots from the experiments.}
  \label{fig:cifar100_resnet18_loss}
\end{figure}

Overall, SUGAR with ELU, SELU, and especially \pms{} delivered the largest gains over the ReLU baseline, whereas LeakyReLU and \nelu{} consistently underperformed (\autoref{fig:cifar100_resnet18_loss}). On CIFAR-10 with a ResNet-18 backbone, \pms{} rose from 76.76 \% to 86.42 \% with SUGAR. VGG-16 showed a similar performance: \pms{} improved the testing accuracy by almost 10 points (78.50 \% → 88.35 \%).

\begin{figure}[ht]
  \centering
  \includegraphics[width=\linewidth]{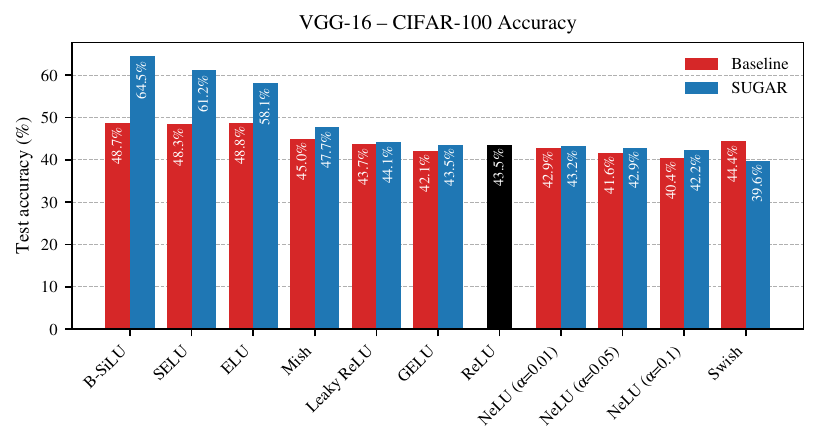}
  \caption{Test accuracy of VGG-16 on CIFAR-100, comparing non-SUGAR (red) and SUGAR (blue) for each activation function. The black bar represents the baseline, where the model is simply trained with ReLU (forward and backward). See \autoref{app:bar_accuracies} for all the accuracy plots from the experiments.}
  \label{fig:vgg16-cifar100-accuracy}
\end{figure}

On CIFAR-100, the superiority of SUGAR \pms{} was even more pronounced: ResNet-18’s accuracy jumped from 48.99  \% to 56.51 \%, and VGG-16’s from 48.73 \% to 64.47 \% (\autoref{fig:vgg16-cifar100-accuracy}). Again, Leaky ReLU and \nelu{} showed negligible or negative gains (e.g. 43.67 \% → 43.41 \% on ResNet-18), indicating that simple linear leakage fails to harness the benefits of SUGAR under this setup. In summary, \pms{} outperforms all other surrogates across architectures and datasets, ELU and SELU provide reliable improvements, and SUGAR does not benefit meaningfully from Leaky ReLU and \nelu{} in this setting.

\subsection{Stability improvements for deep ReLU networks}\label{subsec:DyingReluStability}
To evaluate SUGAR's effectiveness in addressing the dying ReLU problem, we revisited a controlled setting introduced in \cite{lu2020dying}, where a deep and narrow ReLU network with symmetric weight initialization fails to learn due to widespread neuron inactivity. We replicated and extended the original experiments by incorporating SUGAR with \pms{} surrogate gradients, which greatly improved layer activation probabilities and learning outcomes across multiple simple regression tasks. Our results demonstrate that SUGAR enables gradient flow even through inactive neurons, reducing collapse rates and enhancing model expressivity. For an in-depth analysis please refer to \autoref{appendix:toy_task}.
\subsection{Conv2NeXt and Swin Transformer}\label{subsec:conv_swin}
We further investigate SUGAR's potential in state-of-the-art models. In this regard, we have chosen one convolution and one attention based model. Conv2NeXt \cite{feng2022conv2next} was chosen as a convolutional model because it is the same architecture as ConvNeXt \cite{liu2022convnext} but adapted for smaller datasets. Swin Transformer \cite{liu2021Swin} is, on the other hand, a vision transformer model. It is developed for large datasets (i.e. ImageNet$1$k) and the smallest model has $28$M parameters. As we have trained the model on Tiny-ImageNet$200$ \cite{deng2009imagenet}, it is considerably over-parameterized for the dataset. The models were adopted in their original form from \cite{liu2021Swin} and \cite{feng2022conv2next}. The only modification applied was changing the activation function with the corresponding SUGAR implementation.

When applied to Conv2NeXt, SUGAR consistently outperforms the base models that use GELU in both forward and backward pass, as shown in \autoref{tab:swin_conv}. Although our reproduction of the Conv2Next results from \cite{feng2022conv2next} stays lower than the reported $83.84\%$ accuracy with GELU, SUGAR with NeLU exceeds this value. In the case of the Swin Transformer, despite a slight drop in performance for \pms{}, \nelu{} with $\alpha=0.01$ yields a higher accuracy than both base models.

\begin{table}[h]
\centering
\small
\ra{1.3}
\caption{Top-$1$ accuracy ($\%$) of Conv2NeXt and Swin Transformer models, trained on CIFAR-$100$ and Tiny-ImageNet$200$ respectively. Base models do not apply SUGAR. For both models, the original works apply GELU for the activation. Our proposed equations use ReLU in forward pass. NeLU is reported for different $\alpha$ values. \best{Best models} are highlighted.}
\label{tab:swin_conv}
\vspace{\baselineskip}
\begin{tabularx}{\textwidth}{lCCCCCC}
    \toprule
    \textbf{Models} & ReLU(base) & GELU(base) & \pms & NeLU($0.01$) & NeLU($0.05$) & NeLU($0.1$)\\
    \midrule
    \small{Conv2NeXt} & $83.85$\text{\raisebox{0.3ex}{$\scriptstyle\pm$}}$0.2$ & $83.74$\text{\raisebox{0.3ex}{$\scriptstyle\pm$}}$0.2$ & $83.87$\text{\raisebox{0.3ex}{$\scriptstyle\pm$}}$0.2$ & $83.84$\text{\raisebox{0.3ex}{$\scriptstyle\pm$}}$0.1$ & $83.84$\text{\raisebox{0.3ex}{$\scriptstyle\pm$}}$0.3$ & \best{$83.95$\text{\raisebox{0.3ex}{$\scriptstyle\pm$}}$0.2$}\\
    Swin Transformer & $61.25$\text{\raisebox{0.3ex}{$\scriptstyle\pm$}}$0.1$ & $61.39$\text{\raisebox{0.3ex}{$\scriptstyle\pm$}}$0.5$ & $57.15$\text{\raisebox{0.3ex}{$\scriptstyle\pm$}}$0.2$ & \best{$61.48$\text{\raisebox{0.3ex}{$\scriptstyle\pm$}}$0.3$} & $61.24$\text{\raisebox{0.3ex}{$\scriptstyle\pm$}}$0.3$ &  $61.21$\text{\raisebox{0.3ex}{$\scriptstyle\pm$}}$0.2$\\
    \bottomrule
\end{tabularx}
\end{table}
\section{Discussion}
\label{discussion}
In the subsequent analysis, we examine the SUGAR effect by investigating the layer activations for each sample. The results show a clear shift in the activation distribution. Afterwards, the loss landscape of ResNet-18 is analyzed with and without SUGAR. Finally, we explore the potential of treating SUGAR as a form of regularization.

\subsection{SUGAR revives dead neurons}
\label{subsec:activation-profile}
\begin{figure}[ht]
  \centering
  \vspace{-1.cm} 
  \begin{minipage}{0.5\textwidth}
    \includegraphics[width=1.02\linewidth, trim={1cm 0 0 0}, clip]{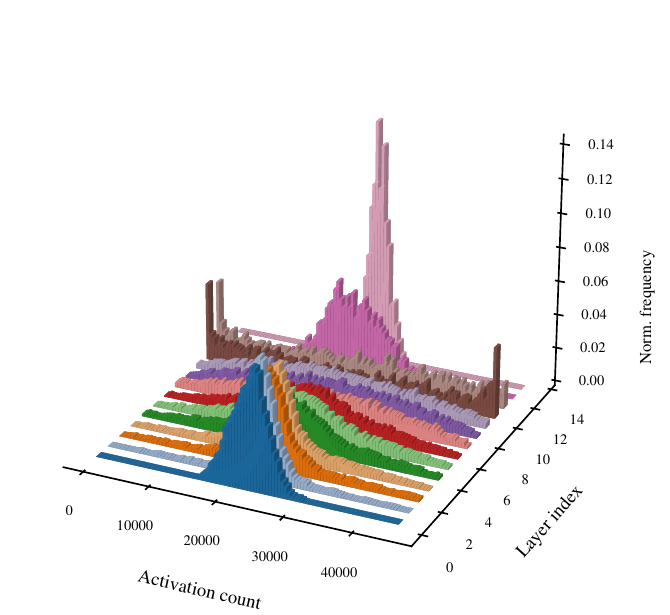}
    \caption*{(a) ReLU}
  \end{minipage}%
  \hfill
  \begin{minipage}{0.5\textwidth}
    \includegraphics[width=1.02\linewidth, trim={1cm 0 0 0}, clip]{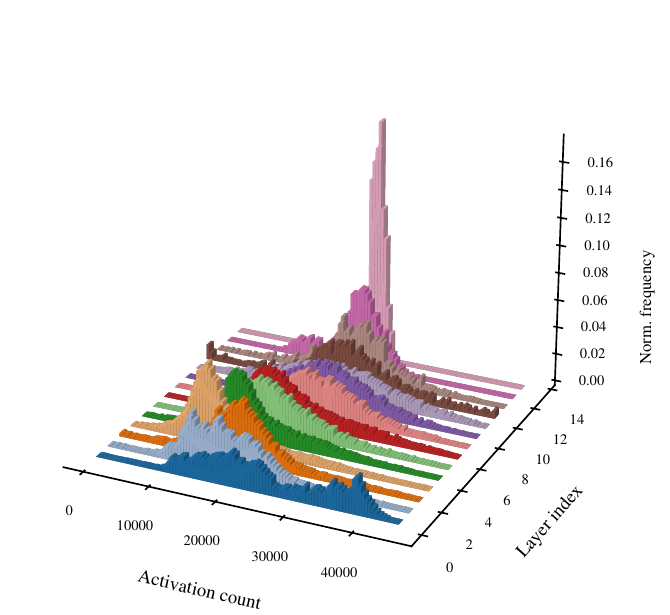}
    \caption*{(b) SUGAR (\pms)}
  \end{minipage}
    \caption{
    Activation profiles of VGG-16 trained on CIFAR-100 after 40 epochs.
    The x-axis shows the activation count per data sample, the y-axis indicates the layer index
    (with the final layer being fully connected), and the z-axis shows the normalized frequency,
    allowing for comparison across layers and activation functions.
  }
  \label{fig:activation_profiles} 
\end{figure}

To shed light on how surrogate gradients affect internal representations, we analyzed the activation profiles of a VGG-16 backbone after 40~training epochs on CIFAR‑100 (see \autoref{fig:activation_profiles}). For each data sample, we recorded how many times a neuron produced an output over the course of an entire training epoch such that the resulting distribution reflects how frequently neurons effectively participate in the forward pass. The frequency at point 0 on the x-axis corresponds to the neurons that are inactive over all samples (i.e. dead neurons).

A first striking difference between the vanilla ReLU baseline and the network trained with SUGAR (\pms) appears in layers~12 and~13. The ReLU model shows a flat distribution in the activation‑count histogram: a sizable group of neurons never fire (dead neurons) while others remain permanently active. In the SUGAR model, the same layers follow an approximately normal distribution centered on moderate activation counts, indicating that bounded surrogate gradients keep gradient flow alive and prevent neurons from becoming functionally inert.

A second difference concerns the shallow part of the network. The first four convolutional layers of the SUGAR model exhibit slightly flatter, right‑skewed distributions whose mode is lower than in the baseline. Hence on average, fewer filters are active for a given image, suggesting that the surrogate‑optimized activation encourages selectivity and reduces redundant feature maps, potentially improving generalization.

Taken together, these observations indicate that bounded surrogate gradients simultaneously mitigate the dead‑neuron problem and promote sparsity where it is most beneficial. The resulting balance—wide early sparsity combined with well‑behaved deep activations—may contribute to the improved generalization reported in \autoref{experiments}. The reduced average activation rate also hints at tangible gains for deployment on resource‑constrained hardware, where memory traffic and multiply‑accumulate counts scale with the number of active neurons.

The present analysis is limited to layer‑wise aggregate statistics. Future work should track the temporal dynamics of activations during training and evaluate whether the same trends generalize to other architectures and datasets. We observed the same activation‑profile patterns on CIFAR‑10 across all tested models. The corresponding plots can be found in \autoref{app:activation_plots}.

\subsection{Loss surface analysis}
\label{subsec:loss-landscape}
\begin{figure}[ht]
    \centering
    \includegraphics[width=5.5in]{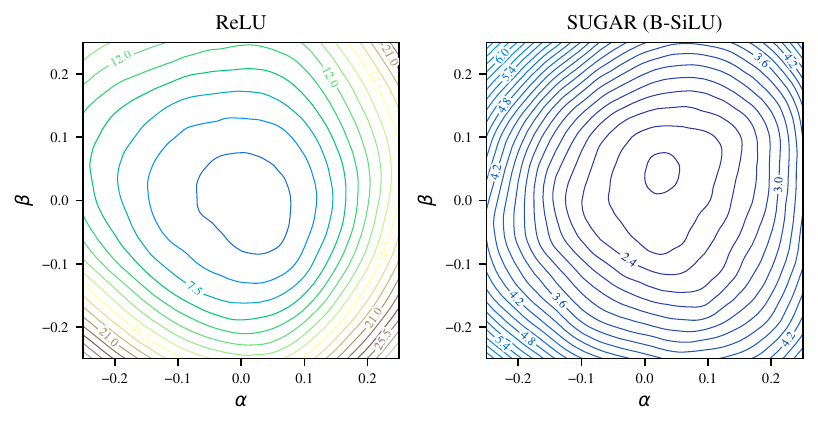}
    \caption{Loss landscapes visualized around the trained solution using different gradient flows. 
    SUGAR (\pms) smooths the optimization surface while retaining the ReLU forward pass, leading to a more stable geometry.}
    \label{fig:loss_landscape}
\end{figure}

To understand how surrogate gradients reshape the optimization geometry, we visualized the loss landscape in the neighborhood of the trained weights after 10~epochs of training a ResNet‑18 on CIFAR‑100 (see \autoref{fig:loss_landscape}). Following the standard two‑direction procedure in \citep{li2018visualizing}, we sampled two random directions, rescaled layer‑wise to match the $\ell_2$‑norm of the corresponding weight tensors (batch‑norm parameters frozen). We evaluated the loss on a $100\times k$ grid spanning $[-0.25,0.25]^2$ and plotted the resulting contours for the vanilla ReLU model and for the same model trained with SUGAR (\pms).

The landscape of the vanilla network exhibits a relatively flat basin at the center but rises steeply toward the edges. At extreme weight perturbations—the corners of the grid—the loss exceeds 25, creating sharp cliffs that can impede optimization. In contrast, the SUGAR landscape is markedly more convex and the loss remains low even at large perturbations. The smoother surface implies better‑conditioned gradients and helps explain the faster convergence we observe during training in \autoref{fig:cifar100_resnet18_loss} and \autoref{app:training_curves}.

\subsection{SUGAR as a regularization technique}
The empirical results in \autoref{subsec:activation-profile} present evidence for a distribution shift in the particular layers. These distributions display sparse activities induced by the ReLU activation function. There is a substantial body of work exploring the relationship between \textit{sparsity} and \textit{generalization} \cite{wen2016learning, gale2019state}. We aim to examine SUGAR from the perspective of regularization. The most widespread and simple regularization is weight decay, which modulates the gradient of the pure cost function towards small weights. SUGAR also regularizes by modulating the gradient of the pure ReLU-based cost function; however, by employing surrogate activation functions, it does so in a more sophisticated and adaptive manner, depending on the activation patterns.

We consider the investigated models in this work. The results in \autoref{subsec:surr} and \autoref{subsec:activation-profile} use models that are not heavily regularized. In this setting, choosing a surrogate function that deviates considerably from ReLU derivative induces harsher regularization and improves the predictive performance. However, in a highly regularized setting such as in \autoref{subsec:conv_swin}, additional regularization has to be applied carefully and nuanced to avoid underfitting. In such a scenario, it is advisable to choose a function (e.g. \nelu) that deviates only slightly from the ReLU's derivative providing gradient flow for pre-activation below the cut-off threshold while modeling the backward characteristics of ReLU as closely as possible. This behavior is reflected in our results: while \pms{} leads to substantial improvements in generalization for VGG-16 and ResNet-18 (see \autoref{subsec:surr}), \nelu{} proves to be more effective in enhancing generalization in the already regularized Conv2NeXt and Swin Transformer models (see \autoref{subsec:conv_swin}).
\section{Conclusion}
\label{sec:conclusion}
This work provides compelling evidence that surrogate gradient learning, originally applied in the spiking neural network domain, can significantly benefit the classical ReLUs in non-spiking deep neural networks. By preserving ReLU in the forward pass while substituting its derivative with a smooth surrogate function during backpropagation, SUGAR enables robust training dynamics and improved generalization, especially in convolutional architectures like VGG-16 and ResNet-18.

Our findings suggest that SUGAR, combined with carefully designed surrogate functions such as B-SiLU and NeLU, offers an elegant solution to the long-standing dying ReLU problem. B-SiLU, in particular, introduces bounded smooth gradients that not only prevent neuron inactivity but also encourage beneficial sparsity patterns. NeLU, on the other hand, offers a more conservative regularization effect through its smooth negative slope, preserving ReLU’s structural simplicity and beneficial properties while improving gradient flow for suppressed activations.

Although the exact influence of NeLU’s negative slope on training dynamics remains an open question, our experiments indicate that it contributes to gradient propagation in strongly regularized models like Conv2NeXt and Swin Transformer. This suggests a nuanced interaction between surrogate gradient shape and model regularization strength, pointing to promising avenues for future exploration.

In conclusion, this work repositions classical ReLU not as a relic, but as a resilient component in the deep learning toolbox. With appropriate gradient handling, ReLU-based networks can match or even outperform modern architectures that rely on more complex activations. 

\subsection{Limitations and future work}
SUGAR’s performance varies notably across different model families. It shows clear benefits in deep, less-regularized networks like VGG-16 and ResNet-18 but is less effective, or even detrimental, in highly regularized architectures such as Conv2NeXt and Swin Transformer, if the surrogate gradient strongly deviates from the ReLU derivative.

The surrogate functions introduced were crafted through empirical intuition and trial-based tuning rather than grounded in formal design principles. Moreover, our results suggest improvements in training dynamics and generalization, but the study does not yet offer formal guarantees on convergence, stability, or generalization bounds. Without a rigorous analytical framework, it is difficult to predict SUGAR’s behavior across different training regimes.

Our evaluation focuses primarily on image classification tasks using datasets like CIFAR-10, CIFAR-100, and Tiny-ImageNet, in addition to few toy problems. It is uncertain how SUGAR would perform in other domains such as natural language processing, reinforcement learning, or time series modeling, where activation dynamics and gradient propagation can differ substantially.

Beyond addressing the current limitations and open questions, future research may assess automatic surrogate function search tailored for specific architectures and datasets. It may also be interesting to consider dynamic surrogates that adapt based on training signals and activation distributions, or in a scheduled manner.

Moreover, given that SUGAR improves sparsity and reduces the activation profiles within the network while applying just simple ReLU, it may prove beneficial for structured pruning, quantization-aware training, or energy-efficient inference in the context of low-footprint models.

{
\small
\bibliographystyle{plain}
\bibliography{references}
}

\newpage
\appendix
\section{A toy example for solving the dying ReLU problem with SUGAR}
\label{appendix:toy_task}
In \cite{lu2020dying}, a deliberate setting for dying ReLUs was created. Under the given configuration, a $10$-layered ReLU network produced constant output due to the dying ReLU problem. The proposed solution is Randomized Asymmetric Initialization (RAI) of the weights such that there are fewer negative activations. In this section, we replicate and augment these experiments with SUGAR and show that SUGAR is able to dramatically reduce the probability of dead activations.

The network in question consists of $10$ feedforward hidden layers with a width of $2$, which only employs ReLU activations. As shown in \cite{lu2020dying}, when a neural network is sufficiently deep relative to its width and initialized with symmetric weights, the initial activation probabilities tend to be near zero. In that case, in more than $90\%$ of trials, the network fails to learn meaningful representations due to widespread irrevocable neuron inactivity. 

To conduct the evaluation, four distinct toy datasets in the range of $[-1.5 , 1.5]$ were adapted from \cite{lu2020dying} as regression tasks. For each function, we drew $3000$ samples from $\textbf{\textit{U}}[-\sqrt{3}, \sqrt{3}]^{d_{in}}$, used as input for the network. $d_{in}$ denotes the input dimension which is $2$ for \autoref{eq:f4} and $1$ for the rest. For each task, $100$ independent runs were carried out. The corresponding equations are as follows:

\begin{align}
    f_1(x) &=|x| \label{eq:f1}\\
    f_2(x) &= x \cdot \sin (5x) \label{eq:f2}\\
    f_3(x) &= \mathbbm{1}_{\{x > 0\}}(x) + 0.2 \sin (5x) \label{eq:f3}\\
    f_4(x_1, x_2) &= \begin{bmatrix} |x_1 + x_2| \\ |x_1 - x_2| \end{bmatrix} \label{eq:f4}
\end{align}

The loss was calculated as the mean squared error over $250$ epochs. Adam \cite{kingma2014adam} was used as an optimizer with a learning rate of $0.005$ and being decreased by $0.1$ in epochs $100$, $150$, $200$, $225$. The minibatch size is $64$ for every target function. As surrogate gradient in this experiment, the derivation of B-SiLU is used.

\begin{figure}[h]
        \centering
        \includegraphics[width=5.5in]{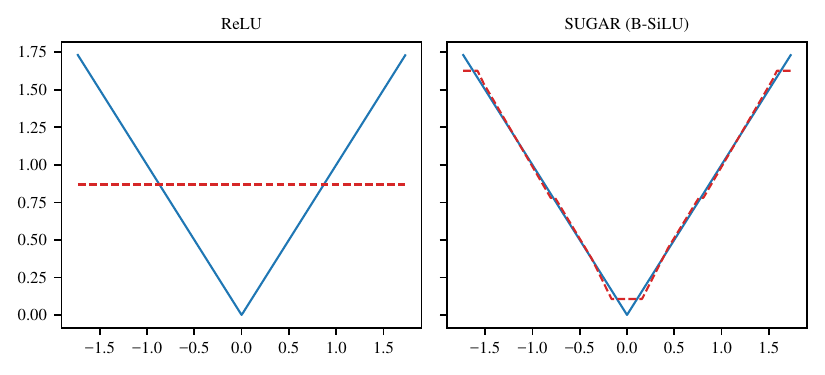}
        \caption{
     Exemplary comparison of predictions. Left plot shows the prediction of the plain ReLU network whereas SUGAR with B-SiLU is applied on the right.
   }
   \label{fig:absfunc_sugar}
\end{figure}

\autoref{fig:absfunc_sugar} illustrates clearly that the ReLU network fails to solve the regression task on \autoref{eq:f1} as has already been shown in \cite{lu2020dying}. Once SUGAR with B-SiLU is applied, the network produces much more activation and is able to solve the regression task.

\begin{figure}[h]
    \centering
    \includegraphics[]{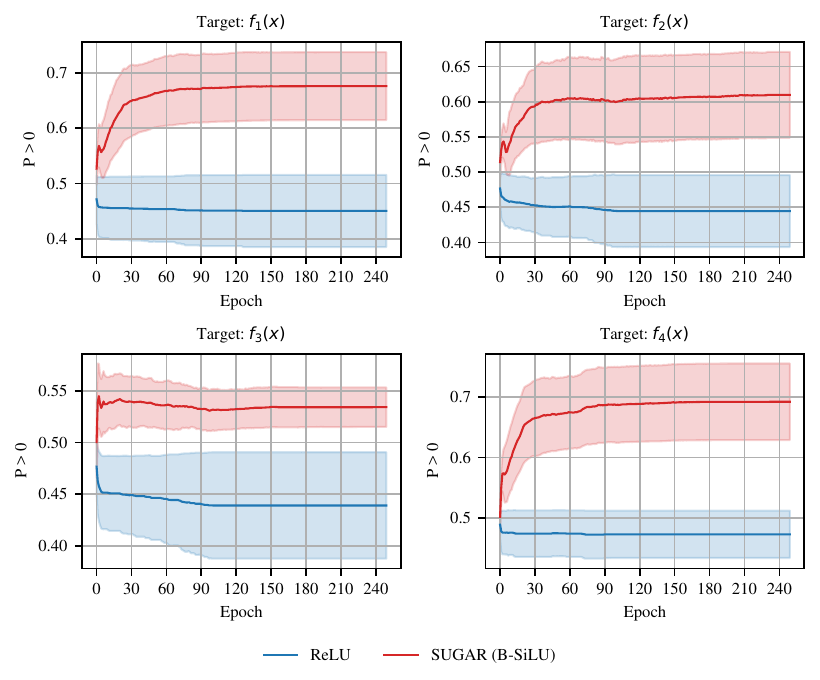}
    \caption{Layer activation probabilities averaged over all runs and layers for each method and task. The plots indicate that SUGAR with \pms{} increase the activations over all layers during training while the use of standard ReLU only leads to a loss or stagnation of layer activity.}
    \label{fig:layer_activation_comp}
\end{figure}

In addition, we tracked the mean layer activation probabilities, as shown in \autoref{fig:layer_activation_comp}, in order to monitor activity within each layer and estimate the number of neurons actively contributing during the forward pass. As a result, we obtained a dynamic view of the network's internal activity, allowing us to assess whether the network utilized its capacity effectively and how sparsity evolved during learning. A model with low activation in early layers might struggle to propagate information, while excessively high activation in later layers may indicate redundancy. Nonetheless, for the sake of clarity, we only visualize the average activities across all layers.

In \cite{lu2020dying}, the number of layers for \autoref{eq:f4} is increased to $20$ from $10$ due to the complexity of the equation. In SUGAR experiments, the layer number is kept at $10$ and the model still generated more activation across layers compared to the results given in \cite{lu2020dying}, as shown in \autoref{tab:collapse_prob}. Using asymmetric positive initialization method, as suggested in \cite{lu2020dying}, reduces the possibility of the neurons becoming inactive. However, once a neuron is inactive, there is no possibility for the model to reactivate the neuron due to the fact that the proposed method only concerns the initialization. For SUGAR, it is still possible as a results of the enabled gradient flow through that particular neuron.

\begin{table}[h]
\centering
\ra{1.2}
\caption{Empirical probabilities in $\%$ from 100 independent simulations of the network to collapse and learn nothing (A), to learn features to some degree (B) and to successfully approximate the target function (C) as indicated in \cite{lu2020dying}. Results annotated with * are taken form \cite{lu2020dying} which conducted $1.000$ independent simulations. For the last three target functions, \cite{lu2020dying} categorized the results only into (A) and not collapsed (D) which is also adapted here.  \best{Best models} are highlighted. The respective loss ranges are decomposed in \autoref{tab:collapse_prob_loss_keys}.}
\label{tab:collapse_prob}
\vspace{\baselineskip}
\begin{tabularx}{\linewidth}{clRRRR}
\toprule
\textbf{Target function} $f_i$ & \textbf{Method} & \textbf{ (A)} & \textbf{ (B)}  & \textbf{(C)} &  (D)\\
\midrule
&Symmetric (He init)* & 93.6 & 4.2 & 2.2 & -\\
\textbf{$f_1(x)$} &RAI* & 40.3 & 22.4 & 37.3 & -\\
&\best{B-SiLU} & \best{15.0} & 8.0 & \best{77.0} & -\\
\midrule
&Symmetric (He init)* & 91.9 & - & - & 8.1\\
\textbf{$f_2(x)$} &RAI* & 29.2 & - & - & 70.8\\
&\best{B-SiLU} & \best{9.0} & - & - & \best{91.0}\\
\midrule
&Symmetric (He init)* & 93.8 & - & - & 6.2\\
\textbf{$f_3(x)$} &RAI* & 32.6 & - & - & 67.4\\
&\best{B-SiLU} & \best{2.0} & - & - & \best{98.0}\\
\midrule
&Symmetric (He init)* & 76.8 & - & - & 23.2\\
\textbf{$f_4(x_1, x_2)$} &RAI* & 9.6 & - & - & 90.4\\
&\best{B-SiLU} & \best{8.0} & - & - & \best{92.0}\\
\end{tabularx}
\end{table}

\begin{table}
\centering
\ra{1.2}
\caption{Each run was assigned to a category by dividing the resulting loss histogram into the respective number of sections. This table provides the loss ranges which were assigned to each category. Note that no loss was observed outside the reported ranges.}
\label{tab:collapse_prob_loss_keys}
\vspace{\baselineskip}
\begin{tabularx}{\linewidth}{CCC}
\toprule
\textbf{Target function} $f_i$ & \textbf{Category} & \textbf{Loss range}\\
\midrule
\textbf{$f_1(x)$} &(A)& $15.5818 \leq \mathcal{L}\leq15.7392$ \\
&(B)& $10.5453 \leq \mathcal{L} \leq 11.0175$\\
&(C)& $0.0003 \leq \mathcal{L} \leq 0.6299$\\
&(D)& - \\
\midrule
\textbf{$f_2(x)$} &(A)& $36.2548 \leq \mathcal{L} \leq 36.6155$\\
&(B)& -\\
&(C)& -\\
&(D)& $0.5452 \leq \mathcal{L} \leq 27.2373$ \\
\midrule
\textbf{$f_3(x)$} &(A)& $19.2409 \leq \mathcal{L} \leq 19.4284$\\
&(B)& -\\
&(C)& -\\
&(D)& $0.6767 \leq \mathcal{L} \leq 1.4268$ \\
\midrule
\textbf{$f_4(x)$} &(A)& $81.0379 \leq \mathcal{L} \leq 81.5630$\\
&(B)& -\\
&(C)& -\\
&(D)& $29.0563 \leq \mathcal{L} \leq 64.7609$ \\
\end{tabularx}
\end{table}
\clearpage
\section{Experimental setup for VGG-16 and ResNet-18}
\label{appendix:experiment_vgg_resnet}
All activation-function experiments on CIFAR-10 and CIFAR-100 were run with identical settings:
\begin{itemize}
  \item \textbf{Batch size:} 128  
  \item \textbf{Validation split:} 10\% of the training set  
  \item \textbf{Data augmentation:} None  
  \item \textbf{Number of workers:} 4  
  \item \textbf{Optimizer:} SGD with learning rate \(0.001\)  
  \item \textbf{LR schedule:} single milestone at epoch 100  
  \item \textbf{Epochs:} 50 for CIFAR-10, 100 for CIFAR-100  
  \item \textbf{Repetitions:} 5 independent runs per configuration with seeds $[1, 10, 20, 25, 42]$
  \item \textbf{Hardware:} NVIDIA RTX A6000 GPUs (CUDA 12.6, Driver 560.35.03)
\end{itemize}
\section{Complete results for VGG-16 and ResNet-18 on CIFAR-10/100}
\label{appdx:vgg_resnet}

In this section the full results of VGG-16 and ResNet-18 are provided. Strikingly, when ReLU and \pms~ join forces, they excel at generalization.

\begin{table}[h]
\centering
\ra{1.2}
\caption{Test accuracy statistics for ResNet-18 and VGG-16 on CIFAR-10. \best{Best model} and the \secondbest{second best model} are highlighted.}
\label{tab:accuracy_sugar_cifar10}
\vspace{\baselineskip}
\begin{tabularx}{\linewidth}{XXCC}
\toprule
\textbf{Model} & Activation & non-SUGAR ($\%$) & SUGAR ($\%$) \\
\midrule
\textbf{ResNet-18} & ELU & \best{${77.91\pm0.49}$} & \secondbest{${83.47\pm0.62}$} 
\\ 
& GELU & $71.57\pm0.52$ & $73.89\pm0.39$ \\
& LeakyReLU & $73.08\pm0.33$ & $73.84\pm0.16$ \\
& Mish & $74.24\pm0.22$ & $79.19\pm0.25$ \\
& \pms & \secondbest{${76.76\pm0.52}$} & \best{${86.42\pm0.33}$} \\ 
& ReLU & $73.22\pm0.42$ & ----- \\
& SELU & $75.90\pm0.63$ & $81.95\pm0.21$ \\
& \nelu($\alpha=0.01$) & $73.01\pm0.32$ & $73.08\pm0.42$ \\
& \nelu($\alpha=0.05$) & $71.01\pm0.55$ & $72.12\pm0.41$ \\
& \nelu($\alpha=0.1$) & $69.42\pm0.34$ & $71.09\pm0.59$ \\
& Swish & $73.91\pm0.49$ & $73.85\pm0.22$ \\
\midrule
\textbf{VGG-16} & ELU &\best{${78.87\pm0.43}$} & $85.58\pm0.16$ \\ 
& GELU & $75.03\pm0.45$ & $75.86\pm0.25$ \\
& LeakyReLU & $75.74\pm0.71$ & $76.04\pm0.26$ \\
& Mish & $76.43\pm0.30$ & $78.98\pm0.33$ \\
& \pms & \secondbest{${78.50\pm0.25}$} & \best{${88.35\pm0.26}$}\\ 
& ReLU & $75.85\pm0.51$ & -----  \\
& SELU & $78.28\pm0.34$ & \secondbest{${86.87\pm0.28}$} \\
& \nelu($\alpha=0.01$) & $75.54\pm0.30$ & $75.88\pm0.23$ \\
& \nelu($\alpha=0.05$) & $74.61\pm0.19$ & $75.50\pm0.33$ \\
& \nelu($\alpha=0.1$) & $73.19\pm0.53$ & $74.75\pm0.17$ \\
& Swish & $75.92\pm0.48$ & $72.81\pm0.48$ \\
\bottomrule
\end{tabularx}
\end{table}

\newpage

\begin{table}[t]
\centering
\ra{1.2}
\caption{Test accuracy statistics for ResNet-18 and VGG-16 on CIFAR-100. \best{Best model} and the \secondbest{second best model} are made salient with the corresponding colors.}
\label{tab:accuracy_sugar_cifar100}
\vspace{\baselineskip}
\begin{tabularx}{\linewidth}{XXCC}
\toprule
\textbf{Model} & Activation & non-SUGAR (\%) & SUGAR (\%) \\
\midrule
\textbf{ResNet-18} & ELU & \best{${49.91\pm0.56}$} & \secondbest{${52.38\pm0.37}$} \\ 
& GELU & $41.55\pm0.43$ & $44.30\pm0.51$ \\
& LeakyReLU & $43.67\pm0.44$ & $43.41\pm0.28$ \\
& Mish & $44.61\pm0.32$ & $48.37\pm0.30$ \\
& \pms & \secondbest{${48.99\pm0.78}$} &  \best{${56.51\pm0.46}$} \\ 
& ReLU & $43.38\pm0.57$ & ----- \\
& SELU & $48.73\pm0.80$ & $52.08\pm0.58$ \\
& \nelu($\alpha=0.01$) & $42.82\pm0.82$ & $43.21\pm0.38$ \\
& \nelu($\alpha=0.05$) & $41.52\pm0.67$ & $42.40\pm0.42$ \\
& \nelu($\alpha=0.1$) & $40.42\pm0.39 $ & $41.94\pm0.17$ \\
& Swish & $43.71\pm0.43$ & $44.05\pm0.26$ \\
\midrule
\textbf{VGG-16} & ELU &  \best{${48.76\pm0.40}$} & $58.09\pm0.40$ \\ 
& GELU & $42.12\pm0.11$ & $43.51\pm0.55$ \\
& LeakyReLU & $43.68\pm0.42$ & $44.11\pm0.30$ \\
& Mish & $44.96\pm0.37$ & $47.74\pm0.12$ \\
& \pms & \secondbest{${48.73\pm0.21}$} & \best{${64.47\pm0.32}$} \\ 
& ReLU & $43.48\pm0.36$ & ----- \\
& SELU & $48.34\pm0.11$ & \secondbest{${61.20\pm0.50}$} \\
& \nelu($\alpha=0.01$) & $42.89\pm0.21$ & $43.18\pm0.28$ \\
& \nelu($\alpha=0.05$) & $41.58\pm0.29$ & $42.85\pm0.20$ \\
& \nelu($\alpha=0.1$) & $40.35\pm0.88$ & $42.19\pm0.47$ \\
& Swish & $44.39\pm0.44$ & $39.59\pm0.31$ \\
\addlinespace
\bottomrule
\end{tabularx}
\end{table}

\section{Activation functions and their derivatives}

\begin{figure}[H]
    \centering
    \includegraphics[width=5.5in]{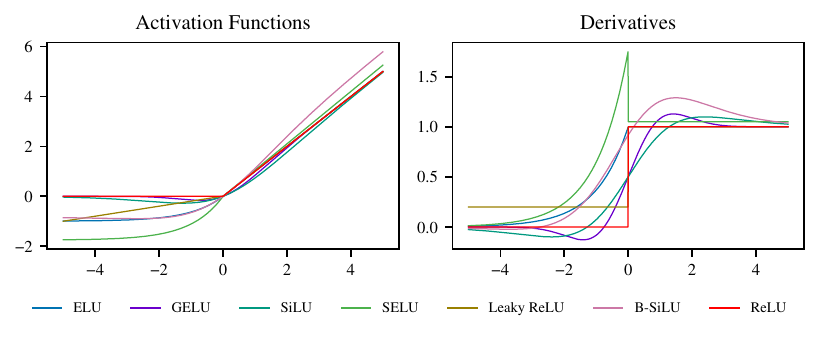}
    \caption{
    Comparison of activation functions and their derivatives used in modern neural networks. The left plot shows the functional forms of ELU, GELU, SiLU, SELU, Leaky ReLU (with slope 0.2), \pms, and ReLU. The right plot shows their respective derivatives. ReLU and \pms{} are overlaid for visual prominence. Notably, non-linear smooth activations exhibit continuous derivatives, while ReLU and its variants introduce discontinuities or sharp transitions. Axis labels denote input $x$ and activation output $f(x)$ or its derivative $f'(x)$.
    }
    \label{fig:act_functions}
\end{figure}

\section{Additional plots from VGG-16 and ResNet-18 experiments}
\label{appendix:additional_plots}
This appendix presents additional plots to provide deeper insight into the convergence behavior, activation profile, and test accuracy observed in the experiments with VGG-16 and ResNet-18.
\subsection{Validation curve plots}
\label{app:training_curves}
\begin{figure}[h!]
    \centering
    \includegraphics[width=5.1in]{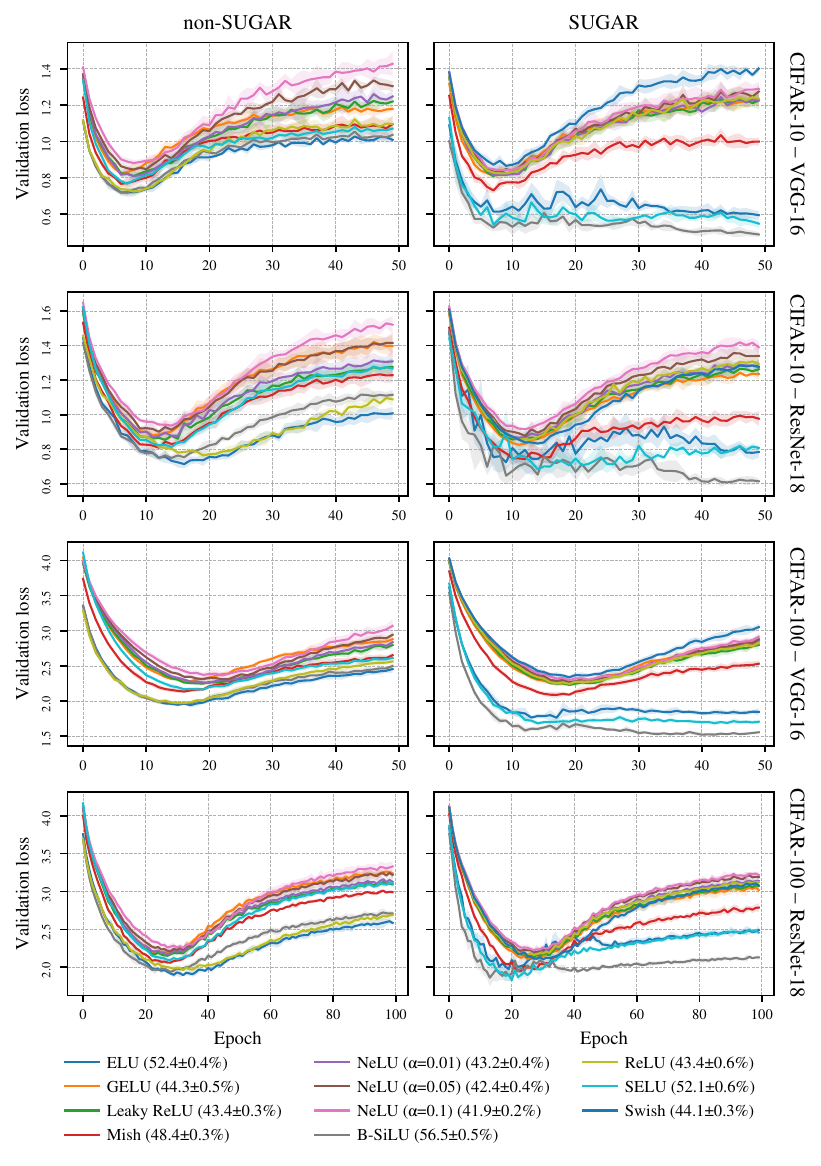}
    \caption{Mean validation‑loss curves for CIFAR‑10/100 on VGG‑16 and ResNet‑18 with confidence intervals.  
    Each experiment pair (row) shares its own log‑scaled \(y\)-axis; experiment titles are displayed vertically on the right‑hand plot in each row.}
    \label{fig:combined_loss_curves}
\end{figure}

\subsection{Activation plots}
\label{app:activation_plots}
\begin{figure}[H]
  \centering
  \vspace{-.8cm}
  \begin{minipage}{0.5\textwidth}
    \includegraphics[width=1.\linewidth]{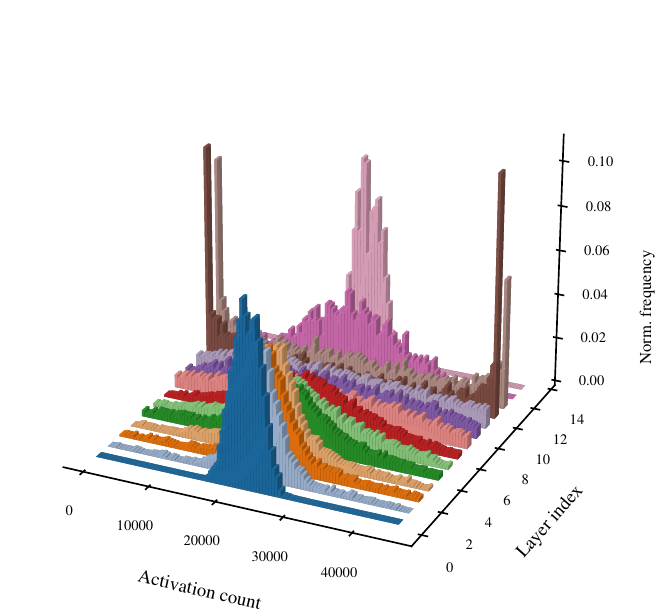}
    \caption*{(a) ReLU (CIFAR-10, VGG-16)}
  \end{minipage}%
  \hfill
  \begin{minipage}{0.5\textwidth}
    \includegraphics[width=1.\linewidth]{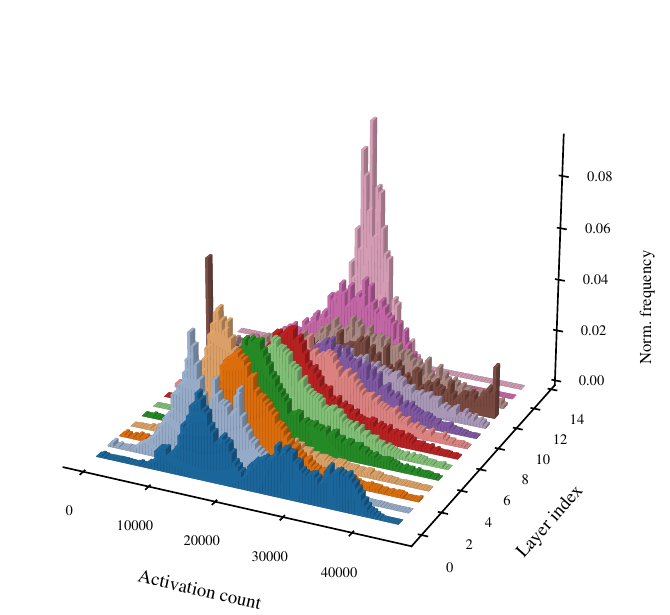}
    \caption*{(b) SUGAR \pms{} (CIFAR-10, VGG-16)}
  \end{minipage}

  \vspace{-.8cm}
  \begin{minipage}{0.5\textwidth}
    \includegraphics[width=1.\linewidth]{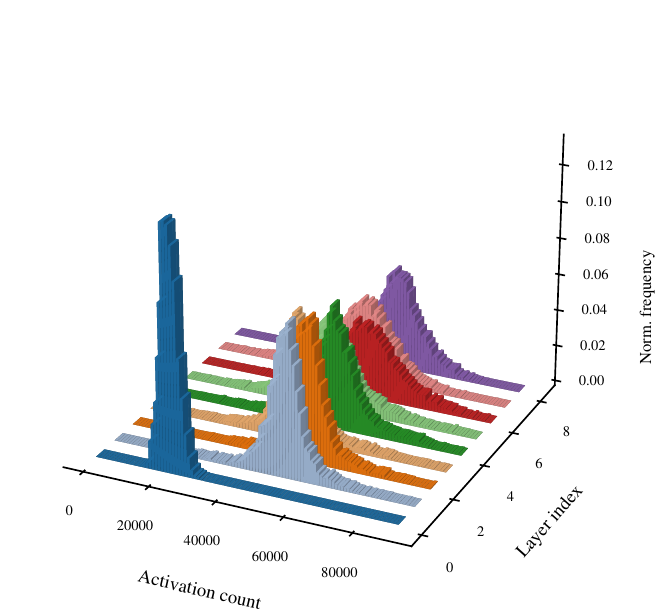}
    \caption*{(c) ReLU (CIFAR-100, ResNet-18)}
  \end{minipage}%
  \hfill
  \begin{minipage}{0.5\textwidth}
    \includegraphics[width=1.\linewidth]{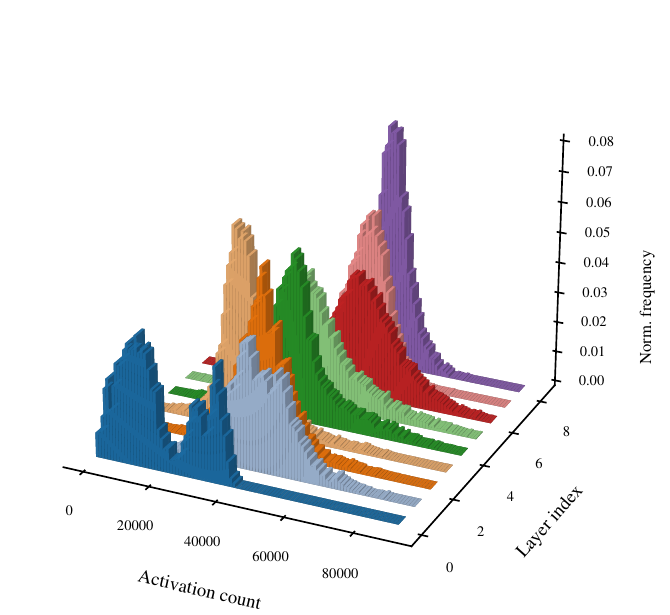}
    \caption*{(d) SUGAR \pms{} (CIFAR-100, ResNet-18)}
  \end{minipage}

  \vspace{-.8cm}
  \begin{minipage}{0.5\textwidth}
    \includegraphics[width=1.\linewidth]{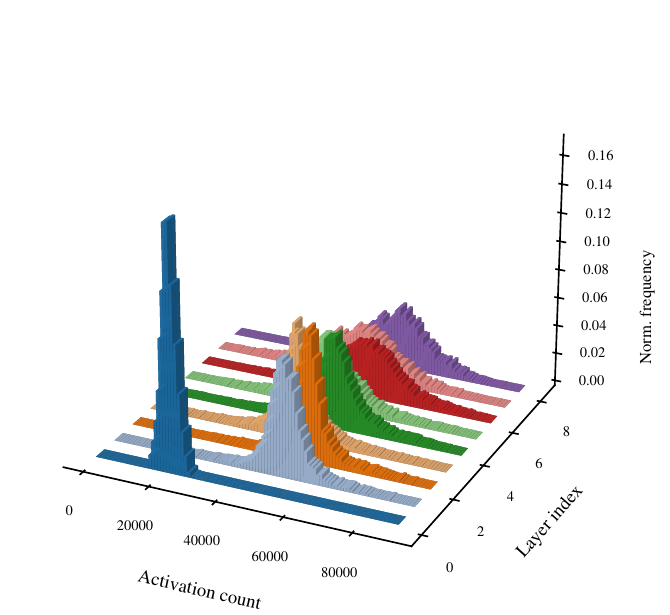}
    \caption*{(e) ReLU (CIFAR-10, ResNet-18)}
  \end{minipage}%
  \hfill
  \begin{minipage}{0.5\textwidth}
    \includegraphics[width=1.\linewidth]{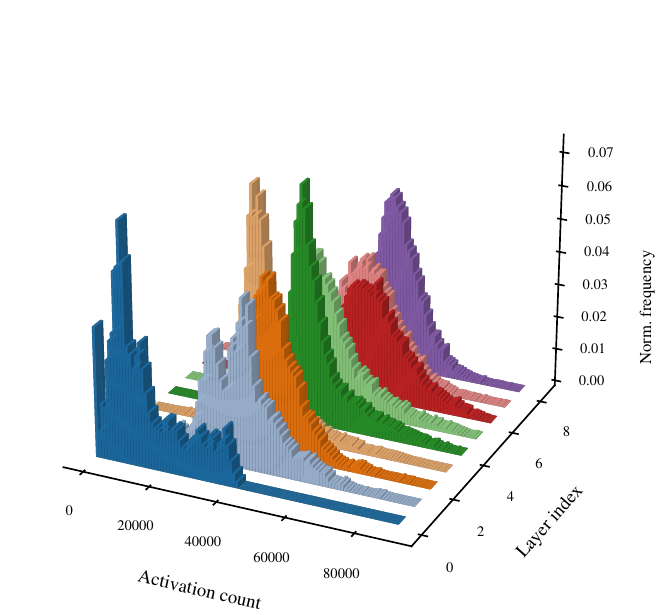}
    \caption*{(f) SUGAR \pms{} (CIFAR-10, ResNet-18)}
  \end{minipage}

  \caption{
    Activation profiles of ReLU and SUGAR \pms{} for VGG-16 and ResNet-18 trained on CIFAR-10 and CIFAR-100.
    The x-axis shows the activation count per data sample, the y-axis indicates the layer index (with the final layer being fully connected),
    and the z-axis shows the normalized frequency, allowing for comparison across layers and activation functions.
  }
  \label{fig:activation_profiles_all}
\end{figure}

\subsection{Bar chart accuracies}\label{app:bar_accuracies}
\begin{figure}[H]
  \centering
  \includegraphics[width=5.5in]{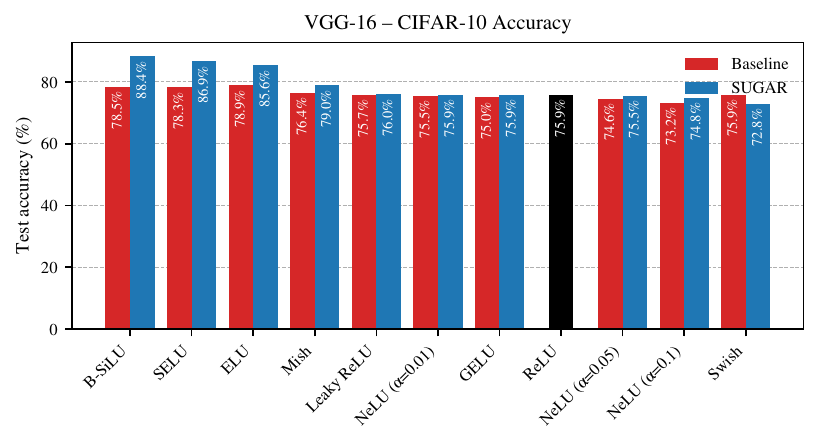}
  \caption{Test accuracy of VGG-16 on CIFAR-10, comparing baseline to SUGAR.}
  \label{fig:vgg16-cifar10-accuracy}
\end{figure}

\begin{figure}[H]
  \centering
  \includegraphics[width=5.5in]{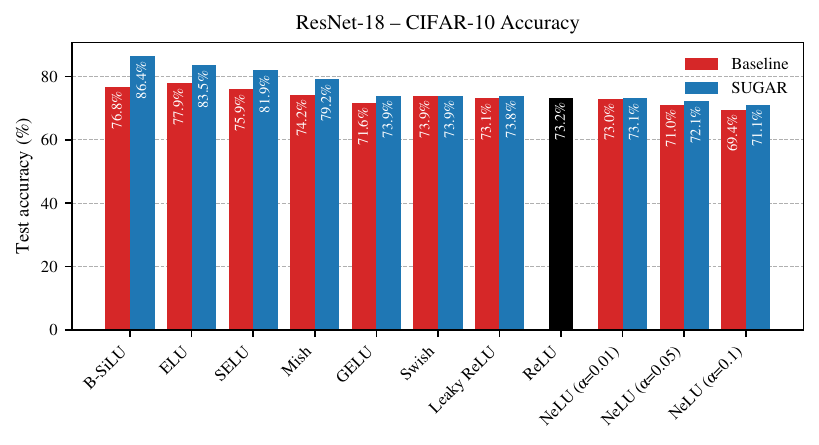}
  \caption{Test accuracy of ResNet-18 on CIFAR-10, comparing baseline to SUGAR.}
  \label{fig:resnet18-cifar10-accuracy}
\end{figure}

\begin{figure}[H]
  \centering
  \includegraphics[width=5.5in]{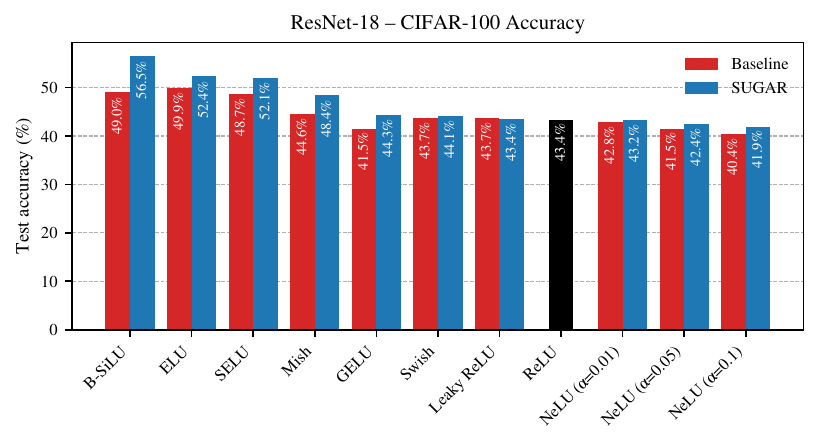}
  \caption{Test accuracy of ResNet-18 on CIFAR-100, comparing baseline to SUGAR.}
  \label{fig:resnet18-cifar100-accuracy}
\end{figure}
\clearpage
\section{Experimental setup for Swin Transformer}
\label{appendix:experiment_swin}
The model is taken from the original work \cite{liu2021Swin}. In this study, the tiny version of Swin Transformer ($28$M parameters) is used. Although our implementation is completely in line with \cite{liu2021Swin}, we provide detailed specifications regarding the training procedure here. Swin Transformer was trained on the Tiny ImageNet dataset with 200 classes.

\begin{itemize}
  \item \textbf{Batch size:} We used a batch size of 200 for training.
  \item \textbf{Validation split:} $5\%$ of the training set

  \item \textbf{Data augmentation:} Several augmentation techniques are employed during training:
  \begin{itemize}
    \item Color jitter with intensity \texttt{0.4}
    \item AutoAugment policy: \texttt{rand-m9-mstd0.5-inc1}
    \item Random erasing with:
    \begin{itemize}
      \item Probability: \texttt{0.25}
      \item Mode: \texttt{pixel}
      \item Count: \texttt{1}
    \end{itemize}
    \item Mixup with alpha \texttt{0.8}
    \item Cutmix with alpha \texttt{1.0}
    \item Mixup and Cutmix applied in \texttt{batch} mode with a switch probability of \texttt{0.5}
  \end{itemize}
  
  \item \textbf{Number of workers:} $8$
  \item \textbf{Optimizer:} \textbf{AdamW} optimizer is used with the following parameters:
  \begin{itemize}
    \item Betas: $(0.9, 0.999)$
    \item Epsilon: \texttt{1e-8}
    \item Weight decay: \texttt{0.05}
  \end{itemize}

  \item \textbf{LR schedule:} A cosine learning rate scheduler was used, with:
  \begin{itemize}
    \item Base learning rate: \texttt{5e-4}, scaled linearly with batch size and number of devices
    \item Warmup learning rate: \texttt{5e-7}
    \item Minimum learning rate: \texttt{5e-6}
    \item Warmup epochs: \texttt{20}
    \item Gradient clipping with max norm: \texttt{5.0}
  \end{itemize}

  \item \textbf{Epochs:} The model was trained for 300 epochs.
  \item \textbf{Repetitions:} 5 independent runs per configuration  with seeds $[1, 10, 20, 25, 42]$
  \item \textbf{Hardware:} NVIDIA GeForce RTX 4090 25GB GPU (CUDA 12.4, Driver 550.144.03)
\end{itemize}

\clearpage
\section{Experimental setup for Conv2NeXt}
\label{appendix:experiment_conv2next}
The model is taken from the original work \cite{feng2022conv2next}. Following the settings in \cite{feng2022conv2next}, the base version of Conv2NeXt ($7$M parameters) is used. Although our implementation is completely in line with \cite{feng2022conv2next}, we provide detailed specifications regarding the training procedure here. \texttt{torch.compile} is applied with SUGAR. Conv2NeXt was trained on the CIFAR-100 dataset.

\begin{itemize}
  \item \textbf{Batch size:} A per-GPU batch size of 200 was used, with gradient accumulation over 4 step, yielding an effective batch size of 800.

  \item \textbf{Validation/training set:} 50000 / 10000 as in default CIFAR-100 setting in torchvision. In accordance with \cite{feng2022conv2next}, no test set was used.

  \item \textbf{Data augmentation:} Following augmentation strategies were deployed:
  \begin{itemize}
    \item AutoAugment: \texttt{rand-m9-mstd0.5-inc1}
    \item Color jitter: \texttt{0.4}
    \item Random erasing with:
    \begin{itemize}
      \item Probability: \texttt{0.25}
      \item Mode: \texttt{pixel}
      \item Count: \texttt{1}
    \end{itemize}
    \item Mixup: \texttt{alpha=0.8}
    \item Cutmix: \texttt{alpha=1.0}
    \item Combined Mixup/Cutmix applied in \texttt{batch} mode with switch probability \texttt{0.5}
  \end{itemize}

  \item \textbf{Number of workers:} $10$

  \item \textbf{Optimizer:} AdamW optimizer is used with the following parameters:
  \begin{itemize}
    \item Learning rate: \texttt{4e-3}
    \item Weight decay: \texttt{0.05}
    \item Epsilon: \texttt{1e-8}
    \item Betas: default (\texttt{(0.9, 0.999)})
  \end{itemize}

  \item \textbf{LR schedule:} A cosine learning rate scheduler was used:
  \begin{itemize}
    \item Initial learning rate: \texttt{4e-3}
    \item Minimum learning rate: \texttt{1e-6}
    \item Warmup period: \texttt{20 epochs}
    \item Weight decay followed a cosine schedule as well
  \end{itemize}

  \item \textbf{Epochs:} Training was performed over 300 epochs.
  \item \textbf{Repetitions:} 5 independent runs per configuration  with seeds $[1, 10, 20, 25, 42]$
  \item \textbf{Hardware:} NVIDIA H100 80GB GPU (CUDA 12.4, Driver 550.127.08)
\end{itemize}
\end{document}